\documentclass{svproc}

\usepackage{url}
\usepackage{hyperref}
\usepackage{subcaption}
\usepackage[pdftex]{graphicx}

\begin{document}
\mainmatter              %
\title{Shape complexity estimation using VAE}
\titlerunning{Comparison of Shape Complexity}  %
\author{Markus Rothgänger\inst{1} \and Andrew Melnik\inst{1} \and
Helge Ritter\inst{1}}
\authorrunning{Markus Rothgänger et al.} %
\tocauthor{Markus Rothgänger, Andrew Melnik, Helge Ritter}
\institute{Bielefeld University, Germany\\
\email{mrothgaenger@techfak.uni-bielefeld.de}
\url{https://github.com/mmrrqq/shape-complexity}
}

\maketitle              %

\begin{abstract}
In this paper, we compare methods for estimating the complexity of two-dimensional shapes and introduce a method that exploits reconstruction loss of Variational Autoencoders with different sizes of latent vectors. Although complexity of a shape is not a well defined attribute, different aspects of it can be estimated. We demonstrate that our methods captures some aspects of shape complexity. Code and training details will be publicly available.

\end{abstract}

\begin{figure}[!h]
  \centering
  \includegraphics[width=0.85\linewidth]{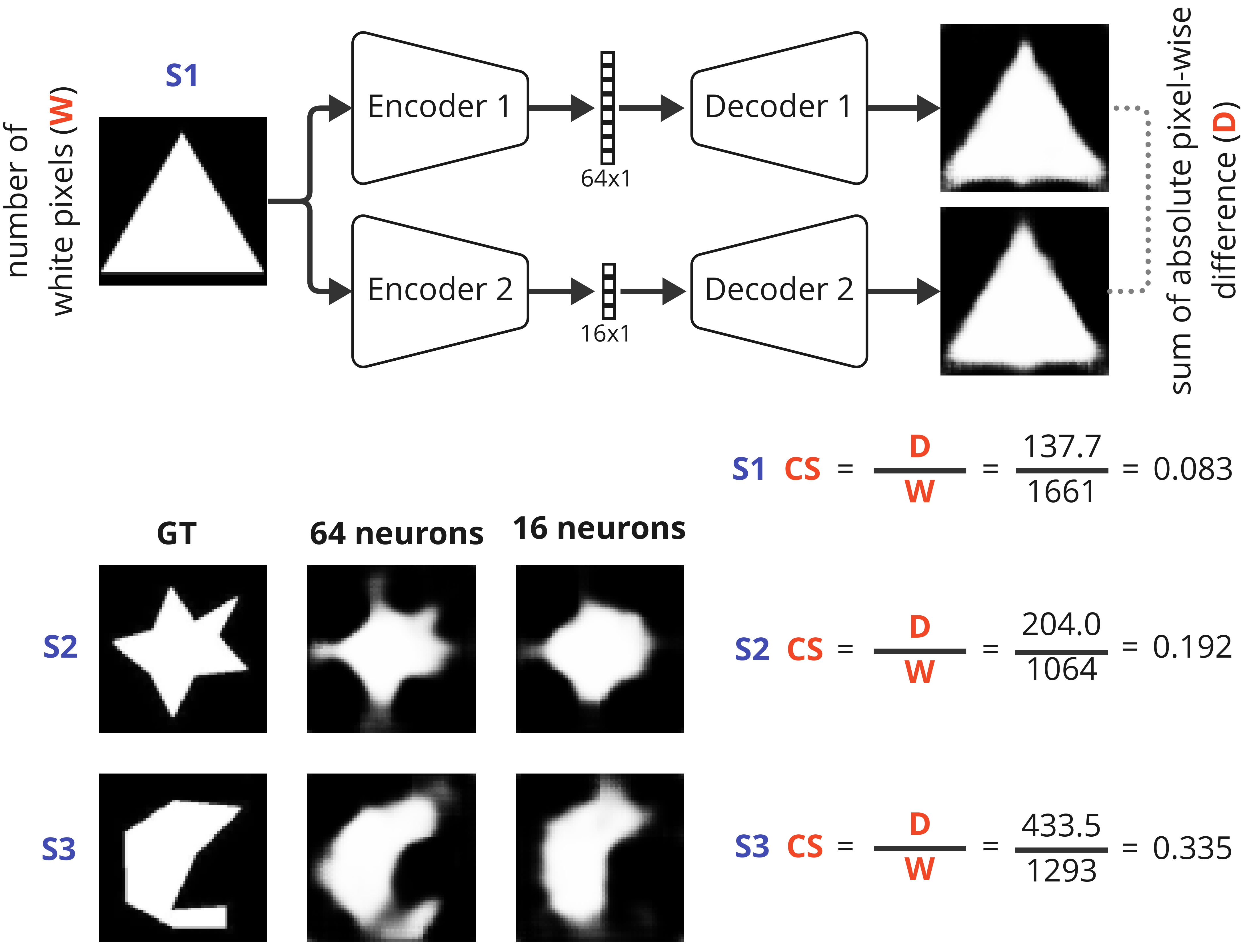}
  \caption{
  Diagram of the architecture and calculation of the Complexity Score (CS) of a shape. The higher the score, the more complex the shape. S1, S2, S3 - Shape 1, Shape 2, Shape 3.}
  \label{fig:architecture}
\end{figure}

\section{Introduction}
The complexity of shapes is not a well defined attribute. However, computer vision systems are faced with the problem of ranking shapes by their complexity.
For the case of two-dimensional shapes, previous definitions for shape complexity used approximations of the Kolmogorov complexity \cite{chen2005} or
the entropy in local features \cite{shape03}. Other work defined strict criteria with circles being the least complex and adding parts that are different from the existing parts of the shape should increase the complexity among others \cite{Chambers2018}\cite{nolte2022stroke}. 
\cite{Chambers2018} applied a range of measures grouped into boundary-based, regional, and skeletal methods \cite{Chambers2018}\cite{Bazazian2022PerceptuallyGQ}. 
These measures are evaluated with a user study \cite{Bazazian2022PerceptuallyGQ} and the authors listed three measures capturing most of the complexity information according to the users: boundary, convexity and skeleton. For humans, the complexity of a shape can be measured by the memory load of remembering the number and arrangement of the composite blocks \cite{melnik2018world}. 
Other methods only consider shapes represented by their outlines \cite{chen2005}\cite{daj} which implies that these methods often miss the inner structure of the shapes. 

Shape complexity estimation exploiting the Fourier analysis was previously conducted in \cite{pedipalp} where the outlines of different biological species were modeled using elliptical Fourier analysis. The method by \cite{costas} considers the inner structure by minimizing the number of ellipses needed to fill the shape while preserving a certain accuracy as an indicator for shape complexity, which however, can get computationally intensive. 
Considering scene like images, the work in \cite{deep} exploits convolutional neural networks to estimate visual complexity in images.

We propose a shape complexity estimation method (see Figure \ref{fig:architecture}) driven by Variational Autoencoders (VAEs). The latent vector of VAE is often used as an information bottleneck \cite{korthals2019jointly}\cite{zai2022transfer}.
We investigated whether complexity measures can be obtained by employing of VAEs with different sized latent vectors for the shape reconstruction step. 

We focus on datasets of two-dimensional binary images and compare complexity orderings obtained with VAEs with more conventional measures based on frequency spectra in Fourier space and an estimation of the Kolmogorov complexity following \cite{chen2005} but using lossless compression. The Kolmogorov complexity and the close relation to data compression is thoroughly examined in \cite{li_introduction_2019}. We further compare our rankings with orderings obtained from human judgements \cite{daj}. We finally propose a measure that combines the VAE with previous measures. Datasets used in this work are described in Section \ref{sec:data} to which follows the introduction of our methods in Sections \ref{sec:vae-method}-\ref{sec:combined-method}.

\section{Methods}
\label{sec:methods}

\subsection{Datasets}
\label{sec:data}
We evaluated and compared methods on three different datasets. First dataset consists of shapes of segments in first-person-view agent from the MineRL dataset \cite{malato2022behavioral}\cite{melnik2021critic}. We also used the MPEG7 shape dataset \cite{mpeg7} that provides a wide range of shape classes from real world examples that are altered in subtle ways within those classes. We further apply the estimation methods to a shape dataset from the work by \cite{daj} as this enables us to compare our results to human-made rankings. Images from these datasets were preprocessed by applying the minimum centered squared bounding box followed by resizing to 64 by 64 pixels.

\subsection{Variational Autoencoder reconstruction measure}
\label{sec:vae-method}
For training the VAE data augmentation is implemented in the form of random horizontal and vertical flipping ($p=0.5$) as well as a random rotation in the range of $\pm 85^\circ$ ($p=0.5$). 

We propose to use the absolute pixel wise difference of the reconstructions obtained from two differently limited VAEs. In our experiments, we limit the latent representations of two VAE to 16 and 64 neurons respectively. Using VAE to estimate complexity arises from the expectation which follows the Kolmogorov complexity that a limited network succeeds at reconstructing simple shapes but generally produces more errors the more complex the shape gets.

We limit the reconstruction capabilities by setting the number of neurons available to represent the latent space for mean and variance of the underlying distributions. For encoding, we use three convolutional layers with ReLU activation and MaxPooling and thereby reduce the $64\times64\times1$ input to $6\times6\times64$. Lastly, two linear layers encode the mean and variance depending on the desired limitation of the network. While decoding the reparameterized latent representation, we first use a linear layer followed by five transposed convolution layers with ReLU activation. The last decoding layer consists of another transposed convolution with sigmoid activation, yielding the same dimensions as the input.

To calculate a single complexity value, our proposed method passes the image through both VAEs and reconstructs from the latent representation. Then, we calculate the absolute pixel wise difference of the two reconstructions divided by the sum of all pixel values in the input image. 
It has to be noted that the reconstructions are not thresholded to bitmasks
and, therefore, yield floating point pixel values from zero for black to one for white.
By its construction, the range of the resulting complexity value is not strictly limited in the interval $[0, 1]$ and might exceed the upper limit, however the value can easily be clipped to one as this would result from a very strong deviation in the reconstructions and therefore suggest a high complexity anyway. To underline this assumption, all absolute pixel wise differences exceeding the sum of white pixel values in the original image (which are always $1$) exceed a complexity value of one.
Both VAEs were trained on the same subset dataset.

\subsection{zlib compression measure}

We approximate the Kolmogorov complexity by calculating the ratio of the byte lengths of the uncompressed image and its compressed counterpart. In theory, a lossless compression of images with homogeneous areas and therefore large chunks of the same data is easier than compressing scattered shape images. As a compression algorithm, we chose to use the zlib implementation of the lossless DEFLATE algorithm \footnote{\url{https://www.rfc-editor.org/rfc/rfc1951}}. For small shapes (with larger homogeneous black areas) with small circumference the compression ratio is not as big as for shapes with large circumference that fill most of their bounding box (i.e., having smaller homogeneous black areas). 
To consider this assumption in our measure, we multiply the compression ratio by $(1-fill\_ratio)$ where $fill\_ratio$ describes the percentage of white pixels in the image. Using this technique we aim to increase the size invariance of our measure.

\subsection{Fourier transform measure}

High spatial frequency is both an indicator for shape complexity, but at the same time it can also indicate image noise \cite{sophistication}. The Fast Fourier Transform (FFT) inevitably lumps both together, reflecting the problem that ``unrecognized complexity" may appear as noise. In some cases, the spatial frequencies of content and noise may be separable, allowing to filter out the noise without affecting the content. Otherwise, a distinction of noise from content is only possible with more detailed models for both, as described in \cite{sophistication}.

While high spatial frequency might not only represent the shape complexity but also image noise as stated above, the FFT measure is based on the assumption that images of complex shapes oftentimes are less homogeneous and therefore contain more high frequency areas. It follows that the average spatial frequency throughout the image gets higher the more complex the shape is.

To find the average frequency of the image, we calculate the two dimensional discrete fourier transform of the image and extract the mean frequency for both dimensions. 
The two frequency means are then combined to a single value using the euclidean norm, yielding a measure for the mean frequency over both dimensions in the range of $0$ to $\sqrt{0.5^2 + 0.5^2}$ cycles per pixel.

Finally, we normalize this measure by dividing it by $\sqrt{0.5^2 + 0.5^2}$. %

\subsection{Combining measures}
\label{sec:combined-method}
As previously stated, we conducted experiments using a combination of the single complexity estimators described above. In line with \cite{Chambers2018}, the combination of measures aims at capturing more than one aspect of complexity.
Although \cite{Chambers2018} suggest that complexity should not be viewed as a one dimensional measure, 
we often wish to have a complexity ranking of shapes.

Therefore, we consider the combination of measures as vectors in an n-dimensional space
where all component values lie within the range $[0, 1]$ where zero suggest low complexity and one high complexity.
To combine their results to a single value, we chose to use the magnitude of the resulting n-dimensional vector.
Although the ranges of the single measures are the same, the values are not necessarily distributed equally over the interval. In simple sorting scenarios a normalization over the min and maximum range of each measure is possible which equalizes the contribution of the measures like already described by \cite{daj}. However is has to be kept in mind that this is not applicable for automated pipelines where the complexity of a single shape needs to be evaluated in a one shot manner.
In the following, we will reference the combination of compression, FFT and VAE measures as the \textit{combined measure}.

\section{Results}
\begin{figure}[!h]
  \centering
  \includegraphics[width=0.55\linewidth]{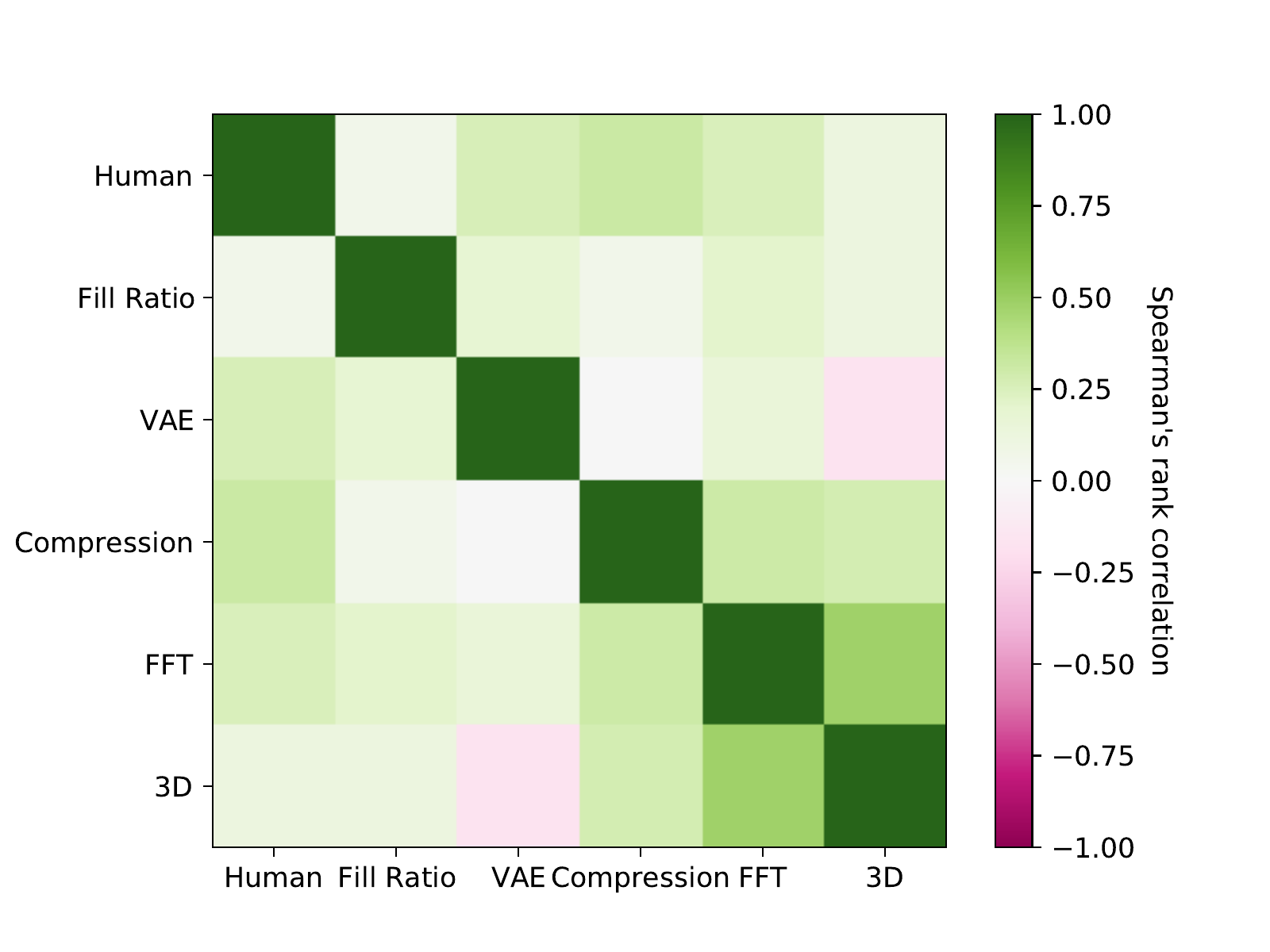}
  \caption{Spearman's rank correlation evaluated on the reference test dataset \cite{daj} with 30 shapes. \textit{Fill Ratio} is the percentage of white pixel in the images, \textit{3D} is the combined measure of VAE, FFT and compression.}
  \label{fig:spearman}
\end{figure}

\begin{figure}[!h]
  \centering
    \includegraphics[clip,width=0.75\textwidth]{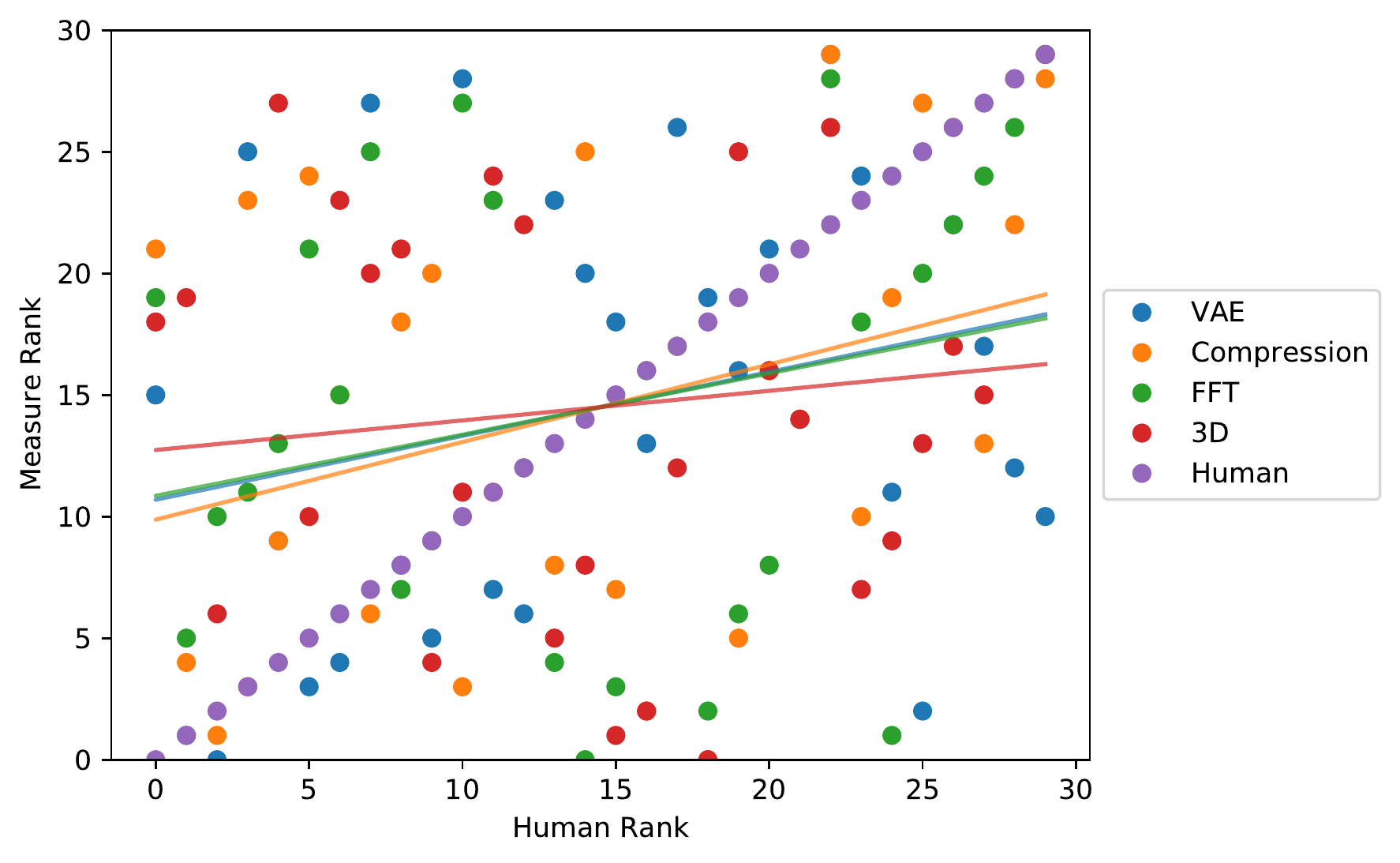}
  \caption{Rankings of measures over human judged rankings including trend lines.}
  \label{fig:distribution-diff}
\end{figure}

A study looking for human judgements of complexity was conducted by Dai et al. in \cite{daj}. We use this dataset to compare the shape-complexity sorting by all methods from Section \ref{sec:methods} and the human-made sorting using the Spearman rank correlation as seen in Figure \ref{fig:spearman}. 
The strongest correlation to the human reference sorting exists for the compression measure. 
The rankings of the measures are plotted against the human-made ranking of the reference dataset in Figure \ref{fig:distribution-diff}. The trendlines allow to define the compression ratio based sorting to be most similar to the human sorting.

\begin{table}[!h]
\centering
\begin{tabular}{l|lll}
            & compression & FFT   & combined    \\ \hline
VAE         & 0.146       & 0.170 & 0.275 \\
compression &             & 0.264 & 0.369 \\
FFT         &             &       & 0.436
\end{tabular}
\vspace{3mm}
\caption{Average Spearman correlation coefficients for rankings of nine randomly sampled shapes (n=2000)}
\label{tab:spearman}
\end{table}

We additionally compare the Spearman's correlation coefficients for rankings of 2000 randomly sampled shape sets of nine shapes from the MPEG-7 dataset \cite{mpeg7} (Table \ref{tab:spearman}).
Considering the single value methods, we observed the strongest correlation between the FFT and the compression based measures. However, with a value of $0.264$ this is a weak correlation. The strongest correlation to the combined measure is given by the FFT based method. These results agree with the previous results and emphasise our informal definition of shape complexity whereas if we consider the combined measure to capture most of the complexity features, there seems to be a slight focus on the frequency distribution of the shapes.

\begin{figure}[h]
  \centering
  \includegraphics[width=0.98\linewidth]{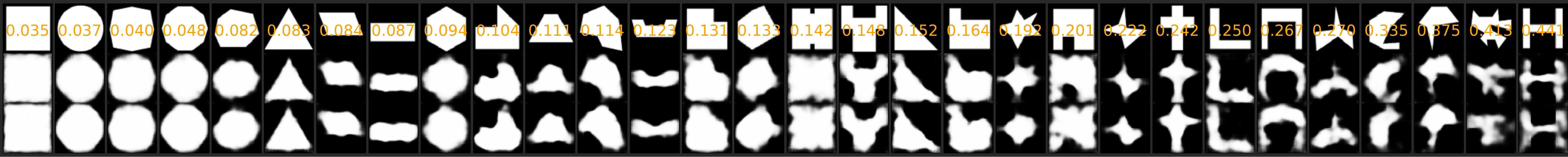}
  \caption{Reconstruction example for the reference dataset \cite{daj}. The first row contains the ground truth images, second row the reconstruction results from the 64 latent neuron VAE, last row the reconstructions by the 16 latent neuron network. The associated complexity value is given in orange.}
  \label{fig:recon-example}
\end{figure}

To visualize the difference in reconstruction capabilities of the VAEs, the reconstructions belonging to the reference dataset \cite{daj} are visualized in Figure \ref{fig:recon-example}. Each cell consists of three images which are the ground truth bitmask image, the 64 latent neuron reconstruction as well as the 16 latent neuron reconstruction (from top to bottom). While large blocks of filled regions are reconstructed with a similar accuracy, the differences are most visible when considering detailed or high frequency regions where the 16 latent neuron network reconstructions are way more blurred.

\begin{figure}[!h]
  \centering
  \begin{subfigure}[c]{\textwidth}
      \begin{minipage}[c]{0.04\textwidth}
        \subcaption{} 
        \label{fig:daj-sort-human}
      \end{minipage}\hfill
      \begin{minipage}[t]{0.96\textwidth}
        \includegraphics[clip,width=\textwidth]{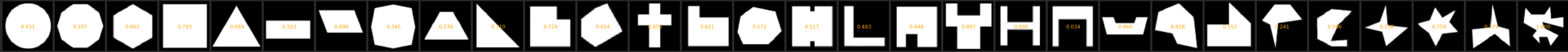}
      \end{minipage}
  \end{subfigure}
  \begin{subfigure}[c]{\textwidth}
      \begin{minipage}[c]{0.04\textwidth}
        \subcaption{} 
      \end{minipage}\hfill
      \begin{minipage}[t]{0.96\textwidth}
        \includegraphics[clip,width=\textwidth]{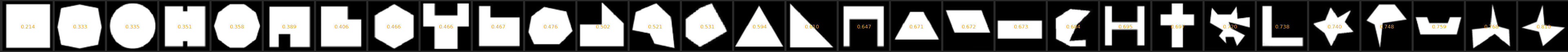}
      \end{minipage}
  \end{subfigure}
  \begin{subfigure}[c]{\textwidth}
      \begin{minipage}[c]{0.04\textwidth}
        \subcaption{} 
      \end{minipage}\hfill
      \begin{minipage}[t]{0.96\textwidth}
        \includegraphics[clip,width=\textwidth]{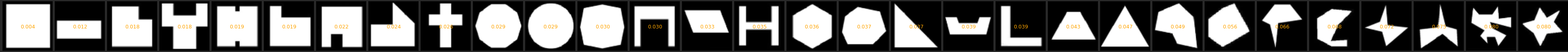}
      \end{minipage}
  \end{subfigure}
  \begin{subfigure}[c]{\textwidth}
      \begin{minipage}[c]{0.04\textwidth}
        \subcaption{} 
      \end{minipage}\hfill
      \begin{minipage}[t]{0.96\textwidth}
        \includegraphics[clip,width=\textwidth]{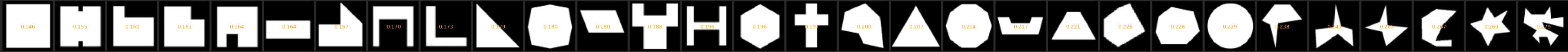}
      \end{minipage}
  \end{subfigure}
  \begin{subfigure}[c]{\textwidth}
      \begin{minipage}[c]{0.04\textwidth}
        \subcaption{} 
      \end{minipage}\hfill
      \begin{minipage}[t]{0.96\textwidth}
        \includegraphics[clip,width=\textwidth]{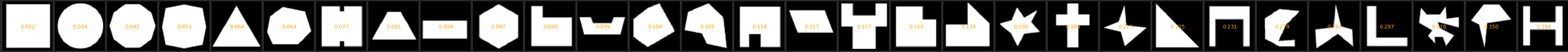}
      \end{minipage}
  \end{subfigure}
  \begin{subfigure}[c]{\textwidth}
      \begin{minipage}[c]{0.04\textwidth}
        \subcaption{} 
      \end{minipage}\hfill
      \begin{minipage}[t]{0.96\textwidth}
        \includegraphics[clip,width=\textwidth]{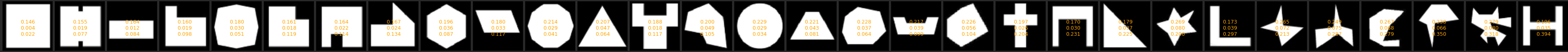}
      \end{minipage}
  \end{subfigure}
  \begin{subfigure}[c]{\textwidth}
      \begin{minipage}[c]{0.04\textwidth}
        \subcaption{} 
      \end{minipage}\hfill
      \begin{minipage}[t]{0.96\textwidth}
        \includegraphics[clip,width=\textwidth]{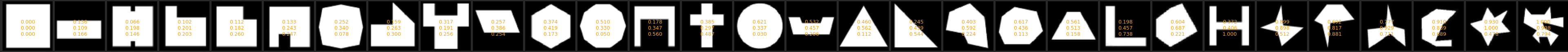}
      \end{minipage}
  \end{subfigure}
  \caption{A comparison of sorting the reference dataset \cite{daj}. The associated complexity value is given in orange text (for the combined measure it is FFT, compression and VAE from top to bottom). (\textit{a} - human reference, \textit{b} - pixel fill percentage, \textit{c} - compression, \textit{d} - FFT, \textit{e} - VAE, \textit{f} - combined, \textit{g} - combined equal contribution)}
  \label{fig:dai-diff}
\end{figure}

\begin{figure}[!h]
  \centering
  \begin{subfigure}[c]{\textwidth}
      \begin{minipage}[c]{0.04\textwidth}
        \subcaption{} 
      \end{minipage}\hfill
      \begin{minipage}[t]{0.96\textwidth}
        \includegraphics[clip,width=\textwidth]{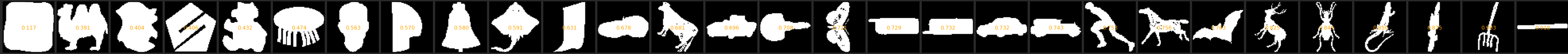}
      \end{minipage}
  \end{subfigure}
  \begin{subfigure}[c]{\textwidth}
      \begin{minipage}[c]{0.04\textwidth}
        \subcaption{} 
      \end{minipage}\hfill
      \begin{minipage}[t]{0.96\textwidth}
        \includegraphics[clip,width=\textwidth]{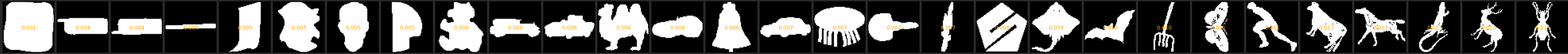}
      \end{minipage}
  \end{subfigure}
  \begin{subfigure}[c]{\textwidth}
      \begin{minipage}[c]{0.04\textwidth}
        \subcaption{} 
      \end{minipage}\hfill
      \begin{minipage}[t]{0.96\textwidth}
        \includegraphics[clip,width=\textwidth]{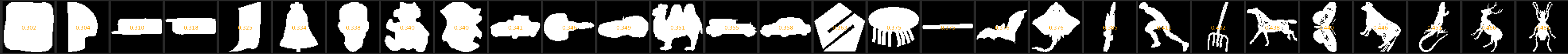}
      \end{minipage}
  \end{subfigure}
  \begin{subfigure}[c]{\textwidth}
      \begin{minipage}[c]{0.04\textwidth}
        \subcaption{} 
      \end{minipage}\hfill
      \begin{minipage}[t]{0.96\textwidth}
        \includegraphics[clip,width=\textwidth]{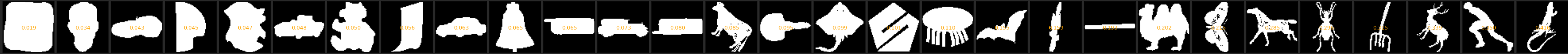}
      \end{minipage}
  \end{subfigure}
  \begin{subfigure}[c]{\textwidth}
      \begin{minipage}[c]{0.04\textwidth}
        \subcaption{} 
      \end{minipage}\hfill
      \begin{minipage}[t]{0.96\textwidth}
        \includegraphics[clip,width=\textwidth]{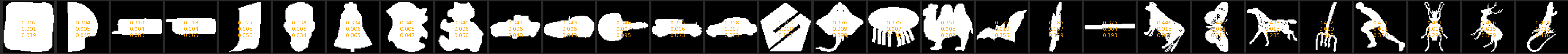}
      \end{minipage}
  \end{subfigure}
  \begin{subfigure}[c]{\textwidth}
      \begin{minipage}[c]{0.04\textwidth}
        \subcaption{} 
      \end{minipage}\hfill
      \begin{minipage}[t]{0.96\textwidth}
        \includegraphics[clip,width=\textwidth]{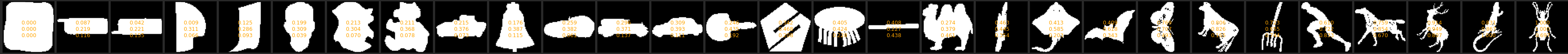}
      \end{minipage}
  \end{subfigure}
  \caption{A comparison of sorting a random subset of the MPEG-7 \cite{mpeg7} dataset using our measures. The associated complexity value is given in orange text (for the combined measure it is FFT, compression and VAE from top to bottom). (\textit{a} - pixel fill percentage, \textit{b} - compression, \textit{c} - FFT, \textit{d} - VAE, \textit{e} - combined, \textit{f} - combined equal contribution)}
  \label{fig:metrics-diff}
\end{figure}

\begin{figure}[!h]
  \centering
  \begin{subfigure}[c]{\textwidth}
      \begin{minipage}[c]{0.04\textwidth}
        \subcaption{} 
      \end{minipage}\hfill
      \begin{minipage}[t]{0.96\textwidth}
        \includegraphics[clip,width=\textwidth]{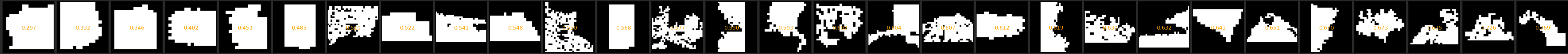}
      \end{minipage}
  \end{subfigure}
  \begin{subfigure}[c]{\textwidth}
      \begin{minipage}[c]{0.04\textwidth}
        \subcaption{} 
      \end{minipage}\hfill
      \begin{minipage}[t]{0.96\textwidth}
        \includegraphics[clip,width=\textwidth]{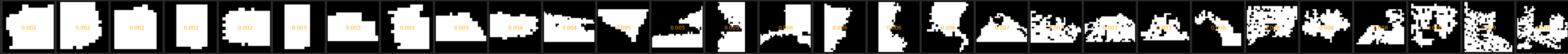}
      \end{minipage}
  \end{subfigure}
  \begin{subfigure}[c]{\textwidth}
      \begin{minipage}[c]{0.04\textwidth}
        \subcaption{} 
      \end{minipage}\hfill
      \begin{minipage}[t]{0.96\textwidth}
        \includegraphics[clip,width=\textwidth]{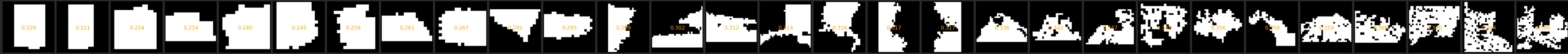}
      \end{minipage}
  \end{subfigure}
  \begin{subfigure}[c]{\textwidth}
      \begin{minipage}[c]{0.04\textwidth}
        \subcaption{} 
      \end{minipage}\hfill
      \begin{minipage}[t]{0.96\textwidth}
        \includegraphics[clip,width=\textwidth]{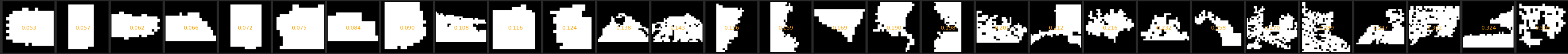}
      \end{minipage}
  \end{subfigure}
  \begin{subfigure}[c]{\textwidth}
      \begin{minipage}[c]{0.04\textwidth}
        \subcaption{} 
      \end{minipage}\hfill
      \begin{minipage}[t]{0.96\textwidth}
        \includegraphics[clip,width=\textwidth]{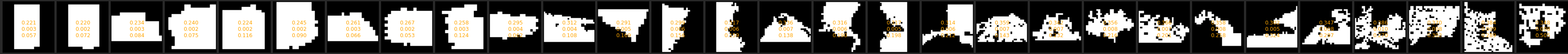}
      \end{minipage}
  \end{subfigure}
  \begin{subfigure}[c]{\textwidth}
      \begin{minipage}[c]{0.04\textwidth}
        \subcaption{} 
      \end{minipage}\hfill
      \begin{minipage}[t]{0.96\textwidth}
        \includegraphics[clip,width=\textwidth]{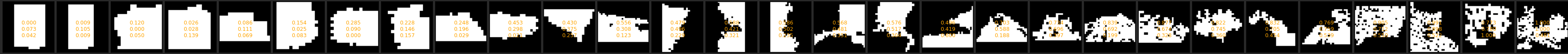}
      \end{minipage}
  \end{subfigure}
  \caption{A comparison for shapes extracted from Minecraft first person views \cite{malato2022behavioral}\cite{melnik2021critic} using our measures. The associated complexity value is given in orange text (for the combined measure it is FFT, compression and VAE from top to bottom). (\textit{a} - pixel fill percentage, \textit{b} - compression, \textit{c} - FFT, \textit{d} - VAE, \textit{e} - combined, \textit{f} - combined equal contribution)}
  \label{fig:metrics-diff-mc}
\end{figure}

\begin{figure}[!h]
  \centering
  \begin{subfigure}[c]{\textwidth}
      \begin{minipage}[c]{0.04\textwidth}
        \subcaption{} 
      \end{minipage}\hfill
      \begin{minipage}[t]{0.96\textwidth}
        \includegraphics[clip,width=\textwidth]{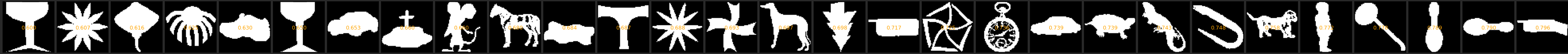}
      \end{minipage}
  \end{subfigure}
  \begin{subfigure}[c]{\textwidth}
      \begin{minipage}[c]{0.04\textwidth}
        \subcaption{} 
      \end{minipage}\hfill
      \begin{minipage}[t]{0.96\textwidth}
        \includegraphics[clip,width=\textwidth]{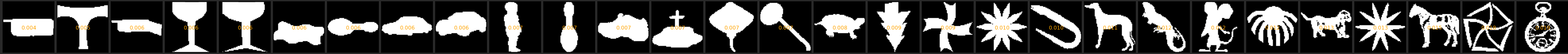}
      \end{minipage}
  \end{subfigure}
  \begin{subfigure}[c]{\textwidth}
      \begin{minipage}[c]{0.04\textwidth}
        \subcaption{} 
      \end{minipage}\hfill
      \begin{minipage}[t]{0.96\textwidth}
        \includegraphics[clip,width=\textwidth]{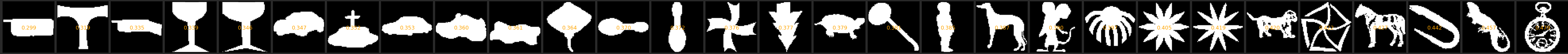}
      \end{minipage}
  \end{subfigure}
  \begin{subfigure}[c]{\textwidth}
      \begin{minipage}[c]{0.04\textwidth}
        \subcaption{} 
      \end{minipage}\hfill
      \begin{minipage}[t]{0.96\textwidth}
        \includegraphics[clip,width=\textwidth]{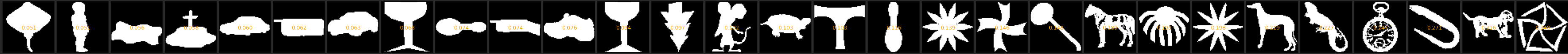}
      \end{minipage}
  \end{subfigure}
  \begin{subfigure}[c]{\textwidth}
      \begin{minipage}[c]{0.04\textwidth}
        \subcaption{} 
      \end{minipage}\hfill
      \begin{minipage}[t]{0.96\textwidth}
        \includegraphics[clip,width=\textwidth]{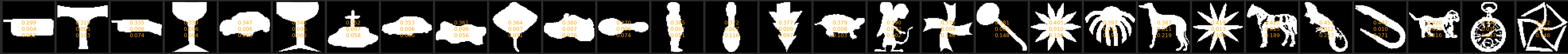}
      \end{minipage}
  \end{subfigure}
  \begin{subfigure}[c]{\textwidth}
      \begin{minipage}[c]{0.04\textwidth}
        \subcaption{} 
      \end{minipage}\hfill
      \begin{minipage}[t]{0.96\textwidth}
        \includegraphics[clip,width=\textwidth]{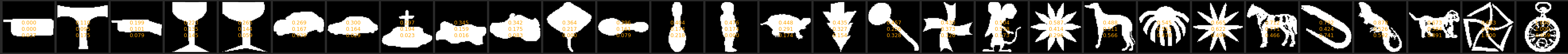}
      \end{minipage}
  \end{subfigure}
  \caption{A comparison of sorting a subset of the MPEG-7 dataset containing only shapes with a white pixel percentage in the interval $(0.6, 0.8)$. The associated complexity value is given in orange text (for the combined measure it is FFT, compression and VAE from top to bottom). (\textit{a} - pixel fill percentage, \textit{b} - compression, \textit{c} - FFT, \textit{d} - VAE, \textit{e} - combined, \textit{f} - combined equal contribution)}
  \label{fig:fill-ratio-diff}
\end{figure}

\begin{figure}[!h]
  \centering
  \includegraphics[width=0.65\linewidth]{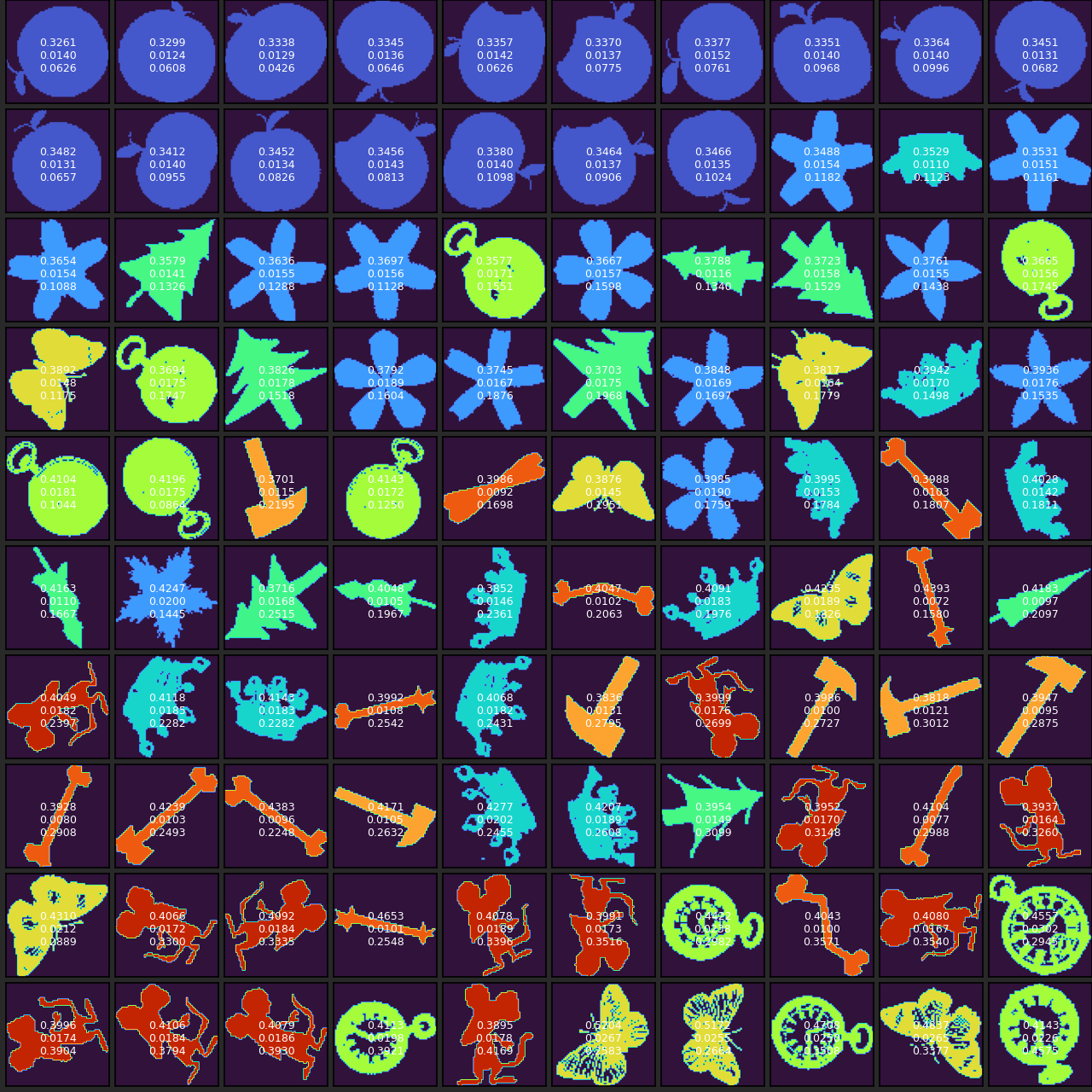}
  \caption{Results of sorting a subset of the MPEG-7 dataset \cite{mpeg7} including ten classes instead of 70 using the normalized combined measure. Shapes belonging to the same class are colorized equally. The associated complexity value for each measue is given as white text for FFT, compression and VAE from top to bottom.}
  \label{fig:3d-grouping}
\end{figure}

Visualization of the sortings for all methods on the reference dataset are given in Figure \ref{fig:dai-diff}.
Applying the measures to a randomly chosen but fixed subset of the MPEG-7 data without augmentation sorts the shapes as depicted in Figure \ref{fig:metrics-diff}. In Addition to the mentioned measures from Section \ref{sec:methods}, we use the percentage of white pixels in the image to analyze the influence of the \textit{fill ratio} on the measures. A visible difference in the sensitivity of the measures can be seen as the compression and FFT measure tend to agree for - what these measures consider to be - the more complex shapes whereas the FFT and the VAE measure agree more on the lower part. In general, if we consider high spectral frequency to be an indicator for high complexity, we notice that large pixelwise differences as well as large compression differences result from high spectral frequency. Due to the values of the VAE measure being more equally distributed over the range $[0, 1]$, naturally, the sorting using the combined measure without min/max normalization of the components contribution does mostly agree with the single VAE measure.
Similar results are visible in Figure \ref{fig:metrics-diff-mc} for shapes extracted via segmentation in Minecraft first person views.

To be able to show the invariance to the white pixel percentage of the images, we show results for a subset of the MPEG-7 dataset \cite{mpeg7} in which we only chose images with $60$ to $80$ percent white pixels in Figure \ref{fig:fill-ratio-diff}. The sorting based on the white pixel percentage in Figure \ref{fig:fill-ratio-diff} \textit{a} deviates strongly from the sortings by our measures and indicates an invariance of the white pixel percentage.

In a similar fashion as introduced in \cite{Chambers2018}, we evaluate the performance of our methods by analyzing how well shapes belonging to the same class of the MPEG-7 dataset are considered to be of the same complexity.

The results in Figure \ref{fig:3d-grouping} are obtained using all three measures combined.
While the classification is not necessarily related to the complexity, some classes of the given dataset are altered in such a way that their level of detail is not changing and the overall shape structure remains the same (apple). For other classes, such as the pocket watch, the detail within the class is changing a lot and we see  simple as well as complex considerations in our measures.
Nevertheless, the sorting seems plausible as shapes with filled regions are considered to be less complex than others.

\section{Discussion, Conclusion, and Future Work}
The informal definition of shape complexity makes it hard to determine the validity of the shown methods. 
To further evaluate the sorting capabilities of different methods a larger user study might be the most fitting evaluation method, similar to the study conducted by \cite{Bazazian2022PerceptuallyGQ}. 

In this paper, we were able to show that traditional image analysis and estimations of the Kolmogorov complexity, including our method using VAEs, are suitable to shape complexity estimation - especially in environments where the computational overhead is required to be limited. While not following strict definitions of complexity, the rankings obtained from the explored methods are correlating to human made rankings and open up the field for further research for computer vision pipelines.

Our proposed method using VAEs shows promising results for multiple datasets while being lightweight in training and application. 

While the normalized combination of all measures utilizes a variety of indicators for complexity, it might be more fitting to use a single measure for complexity estimation in some scenarios where the general structure of the input shapes can be foreseen. 
Rejecting noisy shapes might work best using the FFT or the compression measure while a certain deviation from an expected shape can be evaluated using the VAEs which are trained on the expected shapes.

Although we were able to show that our assumption that limited neural networks are able to reconstruct shapes of low complexity seems to hold in our experiments, the field of reconstructional neural networks yields many more points of interest which might be directed towards shape complexity reconstruction and estimation.
Further investigation of using learning based models might be directed towards the question if it is possible to learn to distinguish complexity and noise while approximating the Kolmogorov complexity, as described by \cite{sophistication}.

\bibliography{sources}
\end{document}